\documentclass[preprint,12pt]{elsarticle}
\usepackage{tabularx}
\usepackage{graphicx}
\usepackage{adjustbox}
\usepackage{setspace}
\usepackage{amsmath}
\usepackage{booktabs}
\usepackage{ltablex}
\usepackage{pdflscape}

\usepackage{amssymb}

\journal{Remote Sensing of Environment}

\begin{document}

\begin{frontmatter}


\doublespacing

\title{Physically Interpretable AlphaEarth Foundation Model Embeddings Enable LLM-Based Land Surface Intelligence}

\author[inst1]{Mashrekur Rahman}
\ead{mashrekur.rahman@dartmouth.edu}

\affiliation[inst1]{organization={Dartmouth College},
            addressline={6025 Baker-Berry Library}, 
            city={Hanover},
            postcode={03755}, 
            state={NH},
            country={USA}}

\begin{abstract}
Satellite foundation models produce dense embeddings whose physical interpretability remains poorly understood, limiting their integration into environmental decision systems. Using 12.1 million samples across the Continental United States (2017--2023), we first present a comprehensive interpretability analysis of Google AlphaEarth's 64-dimensional embeddings against 26 environmental variables spanning climate, vegetation, hydrology, temperature, and terrain. Combining linear, nonlinear, and attention-based methods, we show that individual embedding dimensions map onto specific land surface properties, while the full embedding space reconstructs most environmental variables with high fidelity (12 of 26 variables exceed $R^2 > 0.90$; temperature and elevation approach $R^2 = 0.97$). The strongest dimension-variable relationships converge across all three analytical methods and remain robust under spatial block cross-validation (mean $\Delta R^2 = 0.017$) and temporally stable across all seven study years (mean inter-year correlation $r = 0.963$). Building on these validated interpretations, we then developed a Land Surface Intelligence system that implements retrieval-augmented generation over a FAISS-indexed embedding database of 12.1 million vectors, translating natural language environmental queries into satellite-grounded assessments. An LLM-as-Judge evaluation across 360 query--response cycles, using four LLMs in rotating generator, system, and judge roles, achieved weighted scores of $\mu = 3.74 \pm 0.77$ (scale 1--5), with grounding ($\mu = 3.93$) and coherence ($\mu = 4.25$) as the strongest criteria. Our results demonstrate that satellite foundation model embeddings are physically structured representations that can be operationalized for environmental and geospatial intelligence.
\end{abstract}

\begin{highlights}
\item AlphaEarth satellite foundation model embeddings encode physically interpretable Earth surface properties spanning temperature, vegetation, hydrology, and terrain
\item Embedding-environmental variable relationships demonstrate spatial generalization and temporal stability 
\item Dimension interpretations enable retrieval-augmented generation for natural language environmental queries
\item LLM-as-Judge evaluation with rotating model roles demonstrates the grounding of land intelligence system responses
\end{highlights}


\end{frontmatter}

\doublespacing

\section{Introduction} \label{sec:introduction}

Foundation models are emerging as transformative approaches in earth observation and environmental systems modeling \citep{xiao2025foundation, zhu2024foundations, bodnar2025foundation, data2024multimodal, mai2023opportunities}. These models, trained on massive datasets, learn rich feature representations from unlabeled satellite data that can transfer to diverse downstream tasks. Google's AlphaEarth integrates multi-modal satellite data including optical imagery, SAR, and climatic observations to construct high-dimensional embedding fields for global land surface characterization \citep{brown2025alphaearth, tollefson2025google}.

Despite the growing adoption of satellite foundation models, significant gaps remain in understanding the interpretability of learned embeddings, limiting insights into what physical properties are encoded and how these representations can be leveraged for environmental decision systems. Explainability is particularly challenging in high-dimensional feature spaces where complex, nonlinear relationships are not immediately apparent. While several studies have attempted to explain foundation model behavior through attribution mapping and probing techniques \citep{khan2024transformer, xiao2025foundation, mai2023opportunities}, systematic investigations into whether AlphaEarth embeddings encode physically meaningful environmental features remain limited. Recent applications have demonstrated utility for specific tasks including urban air quality prediction \citep{alvarez2025machine}, agricultural analysis \citep{fang2025leveraging, murakami2025within}, and spatial representation learning \citep{liu2025beyond}. However, comprehensive characterization of the dimension--variable relationships that would enable broader integration into environmental information systems is lacking.

A critical methodological consideration for geospatial models is spatial validation. Spatial autocorrelation can lead to substantial overestimation of generalization capabilities when standard random cross-validation is employed \citep{ploton2020spatial, karasiak2022spatial}. Random validation may conflate memorization of spatially proximate samples with true pattern learning, leading to inflated performance metrics that may not generalize spatially \citep{kattenborn2022spatially, roberts2017cross}. 

Beyond interpretability, the dimension-variable relationships enable novel applications at the intersection of foundation models and natural language interfaces. Retrieval-augmented generation (RAG) systems can leverage embedding similarity search to ground large language model (LLM) responses in actual satellite-derived data. The integration of large language models with geospatial science has emerged as a rapidly growing research frontier, with applications spanning spatial query systems, autonomous GIS workflows, and multi-agent decision support \citep{wang2024gpt, zhang2023geogpt, xu2025agentic, sun2026llm}. Recent frameworks have explored grounding LLM agents in structured earth data for disaster response \citep{chen2025empowering}, natural language retrieval over geospatial databases, and agentic systems unifying multi-modal earth observation data \citep{feng2025earthagent}. Systems leveraging geospatial foundation models require that embeddings encode meaningful, stable relationships with physical variables. Evaluating such systems also presents additional challenges, as traditional metrics may not capture response quality dimensions such as scientific grounding and utility.

In this study, we address these gaps through a multi-step study of AlphaEarth embedding interpretability and its application to land surface intelligence. Using 12.1 million samples across the Continental United States (2017--2023), we systematically characterize relationships between the 64-dimensional embedding space and 26 environmental variables spanning terrain, climate, vegetation, hydrology, and urban development. We test these relationships through spatial and temporal stability analysis, then leverage the validated interpretations to develop a Land Surface Intelligence system that allows natural language queries through retrieval-augmented generation.

Specifically, we address the following research questions:
\begin{enumerate}
    \item Do AlphaEarth embedding dimensions encode physically meaningful environmental features, and can we identify which dimensions correspond to specific land surface properties?
    \item Are embedding-variable relationships robust to spatial validation, and do they remain stable across multiple years?
    \item Can validated dimension interpretations enable retrieval-augmented generation for natural language environmental queries?
    \item How can we evaluate LLM-based geospatial systems for scientific grounding and response quality?
\end{enumerate}

\section{Methods} \label{sec:methods}

\subsection{Study Area and Sampling Design} \label{sec:study_area}

To define our study area, we first created a regular sampling grid across the Continental United States (CONUS), bounded by $125.0^{\circ}$W--$66.5^{\circ}$W longitude and $24.5^{\circ}$N--$49.5^{\circ}$N latitude. Grid points were spaced at $0.025^{\circ}$ intervals (approximately 2.75~km), resulting in approximately 2.34 million locations. We extracted data for all seven available AlphaEarth annual composites (2017--2023), producing approximately 1.73 million samples per year and an aggregated dataset of approximately 12.1 million samples.

\subsection{AlphaEarth Embeddings} \label{sec:ae_embeddings}

We acquired AlphaEarth foundation model embeddings through the Google Earth Engine API \citep{gorelick2017google}. These are 64-dimensional vectors (A00--A63) produced by a geospatial embedding model trained on multi-modal satellite data, including Sentinel-2 optical imagery, Sentinel-1 synthetic aperture radar, Landsat, and ancillary Earth observation sources \citep{brown2025alphaearth}. 

Each annual composite summarizes the full time series of observations for a given location and year into a single embedding vector. We extracted embeddings at 1~km scale with a 500~m point buffer. The distributed 8-bit quantized values were converted to floats for all analyses. Samples with missing embedding values (due to edge effects or data gaps) were excluded.

\subsection{Environmental Variables} \label{sec:env_vars}

We assembled 26 environmental variables spanning seven thematic categories to characterize land surface properties at each sample location (Table~\ref{tab:env_vars}). 

\begin{table}[htbp]
\centering
\caption{Dataset variables and sources. The dataset comprises 64 AlphaEarth embedding dimensions and 26 environmental variables extracted across CONUS (2017--2023), $N \approx 12.1$M samples. Variables marked \textit{static} are temporally invariant; all others are annual composites. Res.\ = native resolution; \textsuperscript{\textit{s}} = resampled to 10\,m at extraction.}
\label{tab:env_vars}
\scriptsize
\resizebox{\textwidth}{!}{%
\begin{tabular}{@{}p{3.2cm}llll@{}}
\toprule
\textbf{Variable} & \textbf{Category} & \textbf{Source} & \textbf{Product / Asset} & \textbf{Res.} \\
\midrule
\multicolumn{5}{l}{\textit{AlphaEarth Embeddings (64 dimensions, A00--A63)}} \\
\addlinespace[2pt]
Embedding (64-D) & --- & Google/DeepMind & \texttt{SATELLITE\_EMBEDDING/V1/ANNUAL} & 10 m \\
\addlinespace[4pt]
\multicolumn{5}{l}{\textit{Terrain}} \\
\addlinespace[2pt]
Elevation (m) & Terrain & USGS & SRTM \texttt{USGS/SRTMGL1\_003} & 30 m\textsuperscript{\textit{s}} \\
Slope ($^{\circ}$) & Terrain & USGS & SRTM (derived) & 30 m\textsuperscript{\textit{s}} \\
Aspect ($^{\circ}$) & Terrain & USGS & SRTM (derived) & 30 m\textsuperscript{\textit{s}} \\
Flow accumulation (log) & Terrain & WWF & HydroSHEDS \texttt{WWF/HydroSHEDS/15ACC} & 500 m\textsuperscript{\textit{s}} \\
\addlinespace[3pt]
\multicolumn{5}{l}{\textit{Soil}} \\
\addlinespace[2pt]
Clay fraction (\%) & Soil & OpenLandMap & SoilGrids \texttt{SOL\_CLAY-WFRACTION} & 250 m\textsuperscript{\textit{s}} \\
Organic carbon & Soil & OpenLandMap & SoilGrids \texttt{SOL\_ORGANIC-CARBON} & 250 m\textsuperscript{\textit{s}} \\
pH (H\textsubscript{2}O) & Soil & OpenLandMap & SoilGrids \texttt{SOL\_PH-H2O} & 250 m\textsuperscript{\textit{s}} \\
Water capacity & Soil & OpenLandMap & SoilGrids \texttt{SOL\_WATERCONTENT-33KPA} & 250 m\textsuperscript{\textit{s}} \\
\addlinespace[3pt]
\multicolumn{5}{l}{\textit{Vegetation}} \\
\addlinespace[2pt]
NDVI (annual mean) & Vegetation & NASA & MODIS \texttt{MOD13A2} 16-day, 1 km & 1 km \\
NDVI (annual max) & Vegetation & NASA & MODIS \texttt{MOD13A2} & 1 km \\
EVI (annual mean) & Vegetation & NASA & MODIS \texttt{MOD13A2} & 1 km \\
LAI (annual mean) & Vegetation & NASA & MODIS \texttt{MOD15A2H} & 500 m \\
Tree cover (\%) & Vegetation & Hansen et al. & \texttt{UMD/hansen/global\_forest\_change} & 30 m\textsuperscript{\textit{s}} \\
Albedo (WSA shortwave) & Radiation & NASA & MODIS \texttt{MCD43A3} & 500 m \\
\addlinespace[3pt]
\multicolumn{5}{l}{\textit{Temperature}} \\
\addlinespace[2pt]
LST daytime ($^{\circ}$C) & Temperature & NASA & MODIS \texttt{MOD11A2} 8-day, 1 km & 1 km \\
LST nighttime ($^{\circ}$C) & Temperature & NASA & MODIS \texttt{MOD11A2} & 1 km \\
Mean air temp. ($^{\circ}$C) & Temperature & PRISM & \texttt{AN81m}, band \texttt{tmean} & 4 km \\
Dew point temp. ($^{\circ}$C)\textsuperscript{$\dagger$} & Temperature & PRISM & \texttt{AN81m}, band \texttt{tdmean} & 4 km \\
\addlinespace[3pt]
\multicolumn{5}{l}{\textit{Climate}} \\
\addlinespace[2pt]
Precip. annual (mm) & Climate & PRISM & \texttt{AN81m}, band \texttt{ppt} (sum) & 4 km \\
Precip. max month (mm) & Climate & PRISM & \texttt{AN81m}, band \texttt{ppt} (max) & 4 km \\
\addlinespace[3pt]
\multicolumn{5}{l}{\textit{Hydrology}} \\
\addlinespace[2pt]
Soil moisture (m$^3$/m$^3$) & Hydrology & ECMWF & ERA5-Land \texttt{MONTHLY\_AGGR}, layer 1 & 11 km \\
Runoff annual (mm) & Hydrology & ECMWF & ERA5-Land (cumulative sum $\times 1000$) & 11 km \\
ET annual (mm) & Hydrology & ECMWF & ERA5-Land (cum. sum, sign-corrected) & 11 km \\
\addlinespace[3pt]
\multicolumn{5}{l}{\textit{Urban / Anthropogenic}} \\
\addlinespace[2pt]
Impervious surface (\%) & Urban & USGS & NLCD \texttt{impervious} band & 30 m\textsuperscript{\textit{s}} \\
Nighttime lights (nW/cm$^2$/sr) & Urban & NOAA & VIIRS DNB \texttt{VCMCFG}, band \texttt{avg\_rad} & 500 m \\
Population density (per km$^2$) & Urban & CIESIN & GPWv4 (2020 est., static) & 1 km\textsuperscript{\textit{s}} \\
\bottomrule
\end{tabular}%
}

\vspace{2pt}
{\scriptsize
\textsuperscript{\textit{s}}Static variable (temporally invariant); all others are annual composites.
\textsuperscript{$\dagger$}Labeled \texttt{temp\_range\_c} in the analysis dataset (legacy column name); derived from PRISM \textbf{tdmean} (mean dew point temperature).
}
\end{table}


\textbf{Terrain:} Elevation and slope were derived from the Shuttle Radar Topography Mission (SRTM) 30~m DEM \citep{farr2007shuttle}. Aspect was computed from the DEM slope surface. Log-transformed flow accumulation was obtained from HydroSHEDS \citep{lehner2008new}.

\textbf{Soil:} Clay fraction (\%), organic carbon content, pH, and available water capacity were obtained from OpenLandMap/SoilGrids \citep{hengl2017soilgrids250m}. These are temporally static and were extracted once per location.

\textbf{Vegetation:} Annual mean and maximum NDVI and mean EVI were calculated from MODIS MOD13A2 16-day composites at 1~km resolution \citep{didan2021modis}. Mean LAI was obtained from MODIS MOD15A2H \citep{myneni2021modis}. Tree cover was taken from the Hansen et al. Global Forest Change baseline \citep{hansen2013high}. Surface albedo was derived from MODIS MCD43A3 \citep{schaaf2002first}.

\textbf{Temperature:} Daytime and nighttime land surface temperature (LST) were obtained from MODIS MOD11A2 8-day composites at 1~km resolution \citep{wan2021modis}. Mean air temperature and dew point temperature were derived from PRISM AN81m monthly normals \citep{daly2008physiographically}.

\textbf{Climate:} Annual precipitation totals and maximum monthly precipitation were computed from PRISM AN81m \citep{daly2008physiographically}.

\textbf{Hydrology:} Volumetric soil moisture (layer 1), annual surface runoff (cumulative, scaled to mm), and annual evapotranspiration (cumulative, sign-corrected, scaled to mm) were obtained from ERA5-Land monthly aggregates \citep{munoz2021era5}.

\textbf{Urban/Anthropogenic:} Mean nighttime lights radiance was obtained from VIIRS DNB monthly composites \citep{elvidge2017viirs}. Population density was taken from GPWv4 at the 2020 estimate \citep{ciesin2018gridded} and held constant across all years.

Terrain and soil variables are temporally static; vegetation, temperature, climate, hydrology, and urban variables were extracted as annual composites for each year. 

\subsection{Interpretability Analysis} \label{sec:analysis}

We applied three complementary methods to characterize the relationships between AlphaEarth embedding dimensions and environmental variables: Spearman rank correlation, Random Forest regression, and a multi-task Transformer model. 

\subsubsection{Spearman Rank Correlation} \label{sec:spearman}

We computed Spearman rank correlation coefficients ($\rho$) between each of the 64 embedding dimensions and each of the 26 environmental variables, producing a $64 \times 26$ correlation matrix. A random subsample of $n = 1{,}000{,}000$ was drawn from the pooled multi-year dataset. We used Spearman's $\rho$ rather than Pearson's $r$ because it captures relationships without assuming linearity. At this sample size, all nonzero correlations achieved $p < 0.001$. For each dimension, the primary corresponding variable was identified as the one with the highest $|\rho|$.

\subsubsection{Random Forest Regression} \label{sec:rf}

We trained separate Random Forest (RF) regressors \citep{breiman2001random} for each of the 26 environmental variables, using all 64 embedding dimensions as predictors. A subsample of $n = 700{,}000$ was drawn from the pooled dataset; after excluding samples with missing target values, effective sample sizes ranged from 309,666 (PRISM-derived variables) to 700,000 (variables with complete spatial coverage). Each model was evaluated using 5-fold cross-validation. Permutation importance was computed on a 100,000-sample subset, producing a $64 \times 26$ importance matrix. For each variable, we recorded the three most important embedding dimensions.

\subsubsection{Multi-Task Transformer} \label{sec:transformer}

We designed a multi-task TabTransformer to jointly predict all 26 environmental variables from the 64-dimensional embedding input. The architecture consists of an input projection layer (scalar to $d_{\text{model}} = 128$ per dimension, yielding 64 tokens), a four-layer Transformer encoder ($h = 8$ attention heads, feed-forward dimension $d_{\text{ff}} = 512$, dropout 0.1), and a shared output MLP ($128 \rightarrow 512 \rightarrow 26$) operating on the mean-pooled token representations.

We trained on $n = 5{,}000{,}000$ samples with batch size 2,048 for 60 epochs on an NVIDIA RTX 5090 GPU (32~GB VRAM) using \texttt{bfloat16} mixed-precision training. Optimization used AdamW \citep{loshchilov2019decoupled} with learning rate $1 \times 10^{-4}$, weight decay 0.01, cosine annealing with 5\% linear warmup, and early stopping.

We extracted two forms of interpretability from the trained model. First, gradient-based importance: for 200,000 samples, we computed the absolute gradient of each output with respect to each input dimension, averaged across samples, and normalized per variable to obtain a $64 \times 26$ importance matrix. Second, attention weights: self-attention weights from the final encoder layer were averaged across heads and across 200,000 forward passes, producing a $64 \times 64$ dimension co-attention matrix.

\subsubsection{Method Convergence} \label{sec:convergence}

For each dimension, we compared primary variable assignments across the three methods. A dimension was considered concordant if at least two of the three methods agreed on the primary associated variable. Across all $64 \times 26 = 1{,}664$ dimension--variable pairs, the Pearson correlation between $|\rho|$ (Spearman) and RF permutation importance was $r = 0.45$, indicating moderate convergence between the linear and nonlinear characterizations.

\subsection{Validation} \label{sec:validation}

\subsubsection{Spatial Block Cross-Validation} \label{sec:spatial_cv}

Random cross-validation can overestimate performance in geospatial settings because spatially proximate samples share information through spatial autocorrelation \citep{ploton2020spatial, roberts2017cross}. To assess spatial generalization, we partitioned CONUS into $2^{\circ} \times 2^{\circ}$ blocks and assigned blocks to five folds using grouped $k$-fold splitting, ensuring all samples within a block appear in the same fold. We applied spatial cross-validation to both the RF and Transformer models and computed the generalization gap as $\Delta R^2 = R^2_{\text{random}} - R^2_{\text{spatial}}$ for each variable.

\subsubsection{Temporal Stability} \label{sec:temporal}

We assessed whether embedding--variable relationships remain stable across years by computing year-specific Spearman correlation profiles. For each year (2017--2023), we independently drew $n = 300{,}000$ samples and computed the full $64 \times 26$ correlation matrix. This produced seven annual correlation profiles per dimension, each a 26-element vector of $\rho$ values.

Temporal stability was quantified for each dimension as the mean pairwise Pearson correlation between its annual profiles across all $\binom{7}{2} = 21$ year-pairs. A value near 1.0 indicates that the dimension's relationships with environmental variables are consistent over time, while low values would indicate sensitivity to interannual variability.

\subsection{Dimension Dictionary} \label{sec:dim_dict}

We compiled the outputs of all three methods into a dimension dictionary: a structured lookup table recording each dimension's primary Spearman variable ($\max |\rho|$), primary RF variable ($\max$ importance), primary Transformer variable ($\max$ gradient importance), associated correlation strengths and importance scores, and thematic category assignments. The dictionary also records two-way (Spearman--RF) and three-way concordance flags. This dictionary serves as the interpretive backbone of the Land Surface Intelligence system, translating raw embedding values into environmental meaning at query time.

\subsection{Land Surface Intelligence System} \label{sec:system}

\begin{figure*}[!ht]
\centering
\includegraphics[width=\textwidth, trim=0 30mm 0 20mm, clip]{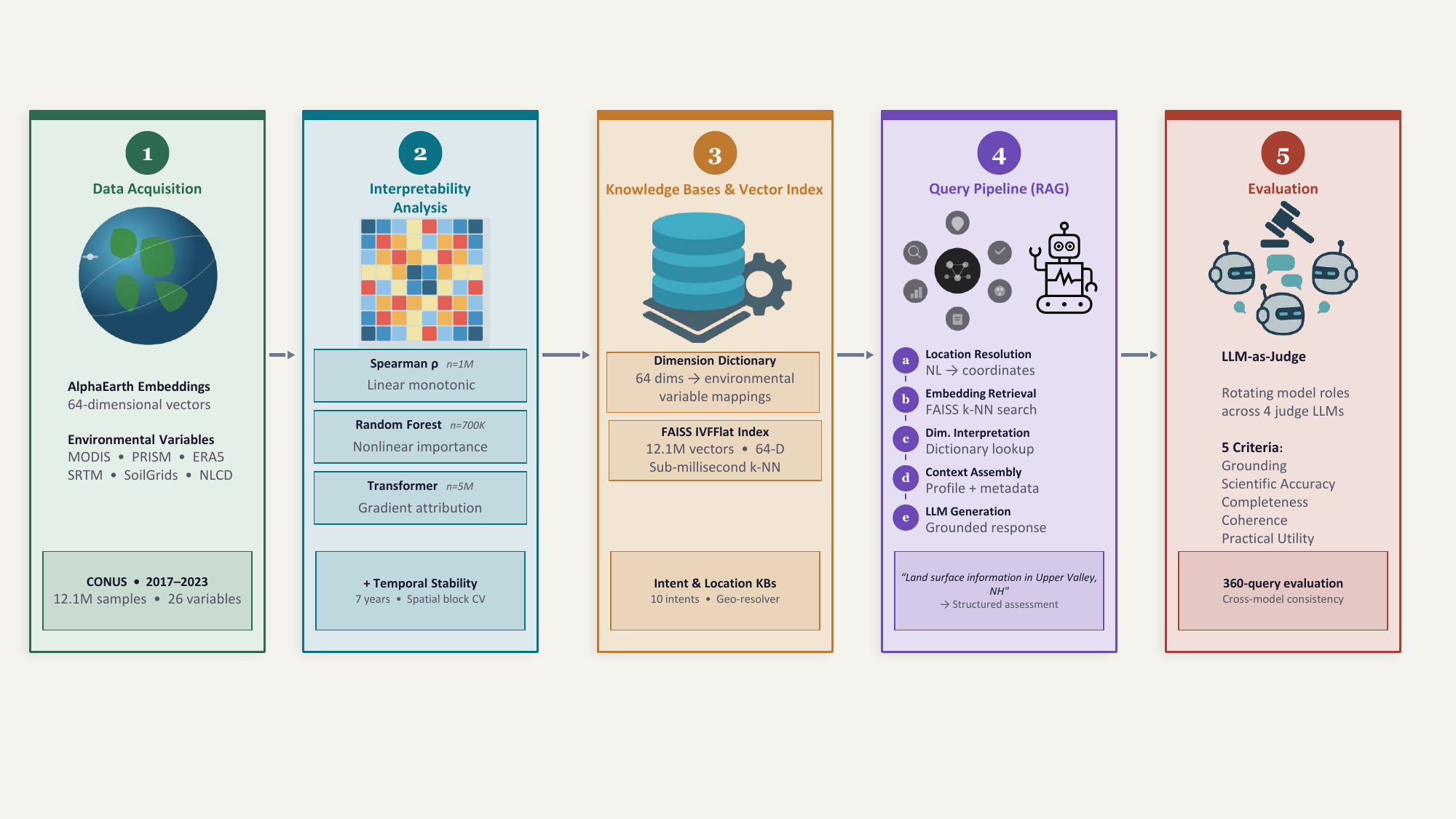}
\caption{Land Surface Intelligence system architecture. The pipeline begins with a natural language query that is resolved to geographic coordinates and a target year. The corresponding AlphaEarth embedding vector and environmental variables are retrieved from the FAISS-indexed database of 12.1 million samples. Each embedding dimension is interpreted using the dimension dictionary compiled from the Spearman, Random Forest, and Transformer analyses. The query is classified into one of ten intent categories, and the $k$ most similar locations are retrieved via nearest-neighbor search. The structured context document, containing location data, dimension interpretations, and similar-location metadata, is provided to a large language model through retrieval-augmented generation to produce a grounded environmental assessment.}
\label{fig:system_architecture}
\end{figure*}

We developed a Land Surface Intelligence system that implements retrieval-augmented generation (RAG) over the AlphaEarth embedding space, enabling natural language environmental queries grounded in satellite-derived data (Figure~\ref{fig:system_architecture}).

\subsubsection{FAISS Vector Index} \label{sec:faiss}

We indexed the full set of approximately 12.1 million embedding vectors using Facebook AI Similarity Search (FAISS) \citep{johnson2019billion}. We used \texttt{IndexIVFFlat} with $n_{\text{list}} = 3{,}500$ clusters (approximately $\sqrt{N}$) and search-time $n_{\text{probe}} = 64$.  Each vector is linked to a metadata record containing coordinates, year, and all 26 environmental variables, stored in a Parquet file. Query latency is sub-millisecond for $k$-nearest-neighbor retrieval.

Proximity in AlphaEarth embedding space corresponds to similarity in physical land surface characteristics. A query location in one geographic region can therefore retrieve physically analogous locations across CONUS, enabling comparative environmental analysis without explicit feature matching.

\subsubsection{Query Pipeline} \label{sec:query_pipeline}

We developed a location-first query pipeline that translates natural language queries into structured environmental profiles through five sequential stages. In the first stage, location resolution, natural language location references (e.g., ``Upper Valley, NH'') are resolved to geographic coordinates through a hierarchical matching process. Temporal references corresponding to the available data years (2017--2023) are extracted when present; otherwise, the system defaults to the most recent available year.


The second stage retrieves the embedding vector for the resolved location. Given coordinates $(lon, lat)$, we identify the nearest indexed point using Euclidean distance in geographic space. The corresponding 64-dimensional embedding vector and associated environmental variables are retrieved from the metadata store. 


In the third stage, each embedding dimension is interpreted using the dimension dictionary constructed from our interpretability analysis (Section~\ref{sec:dim_dict}). For dimensions exhibiting correlations $|\rho| > 0.5$ with environmental variables, we generate structured interpretations linking dimension values to environmental meaning. For example, a high value in dimension A57 combined with its positive correlation with precipitation ($\rho = +0.78$) indicates a location within a wet climate regime.

The fourth stage classifies user queries into one of ten intent categories: flood risk, drought vulnerability, vegetation health, agricultural suitability, climate characterization, terrain analysis, hydrology, urban development, location comparison, and general profile. Classification employs keyword matching against a predefined taxonomy, where each intent category is associated with a set of trigger keywords and a prioritized list of relevant environmental variables. Intent classification helps us navigate which variables and dimension interpretations are emphasized in the response.

The fifth stage executes similarity search to identify the $k$ most similar locations (default $k = 10$) via the FAISS index. 

\subsubsection{LLM Integration} \label{sec:llm_integration}

The outputs of the retrieval pipeline are integrated with a large language model through retrieval-augmented generation (RAG). We constructed a system prompt that encodes the validated dimension--variable relationships established through our interpretability analysis. This grounding ensures that LLM responses reflect empirically validated interpretations. The system prompt specifies key relationships (e.g., ``A57 $\rightarrow$ Precipitation ($\rho = +0.78$): Higher values indicate wetter climates'') and instructs the model to reference actual retrieved data in its responses.

For each user query, we construct a context document containing the resolved location and coordinates, retrieved environmental variable values with units and interpretations, key dimension interpretations ranked by absolute embedding value, a synthesis section tailored to the detected intent, and metadata for similar locations. The LLM receives this structured context along with the user's original query and generates a natural language response that synthesizes the satellite-derived information into an actionable environmental assessment.

We implemented the system using the Dartmouth Chat API \citep{dartmouth_chat} with access to multiple LLM backends. For the evaluation experiments, we selected four free-tier models to maximize query budget while maintaining response quality: GPT-OSS-120B (a reasoning-focused open-source model), Llama-3.2-11B-Vision-Instruct \citep{grattafiori2024llama3}, Gemma-3-27B-IT \citep{gemma2025gemma3}, and Qwen3-VL-32B-Instruct \citep{yang2025qwen3}. These models span different architectural families and training approaches, providing diversity for cross-model evaluation.

\subsection{System Evaluation} \label{sec:evaluation}

Evaluating LLM-based geospatial systems presents challenges beyond traditional retrieval metrics, as response quality encompasses scientific grounding, factual accuracy, and practical utility for decision support. We adopted an LLM-as-Judge framework \citep{zheng2023judging} with rotating model roles to mitigate single-model bias and provide multi-perspective assessment of system performance.

\subsubsection{Evaluation Design} \label{sec:eval_design}

We designed a cross-model evaluation experiment structured around three functional roles: Query Generator, System Under Test, and Judge. The Query Generator produces natural language queries about randomly sampled CONUS locations. The System Under Test processes these queries through our Land Surface Intelligence pipeline with a specific LLM backend. The Judge evaluates the quality of system responses against defined criteria.

Four free-tier LLMs were rotated through these roles according to a design constraint: the generator and judge models must differ from the system model to prevent self-evaluation bias. This constraint yields 12 distinct configurations (4 system models $\times$ 3 generator/judge model combinations). For each configuration, we generated 30 queries, producing 360 total evaluation cycles across the experiment.

Query locations were sampled uniformly within CONUS bounds ($125.0^{\circ}$W--$66.5^{\circ}$W, $24.5^{\circ}$N--$49.5^{\circ}$N). Each query was assigned one of ten intent types through round-robin allocation to ensure balanced coverage across use cases. Query text was generated using templates specific to each intent type with natural language variation (e.g., ``What is the flood risk for [location]?'' or ``Assess the drought vulnerability of [location]'').

\subsubsection{Evaluation Criteria} \label{sec:eval_criteria}

System responses were evaluated on five criteria with associated weights reflecting their importance for scientific applications. Grounding (weight = 0.25) assesses whether the response references actual embedding data and environmental variables with correct interpretations, distinguishing data-driven responses from generic or speculative statements. Scientific accuracy (weight = 0.25) evaluates whether interpretations are consistent with the validated dimension--variable relationships from our analysis, such as correctly interpreting A48's positive correlation with EVI or A57's association with precipitation. Completeness (weight = 0.20) measures whether the response fully addresses the user's query with relevant environmental categories appropriate to the detected intent. Coherence (weight = 0.15) assesses whether the response is well-structured, clear, and logically organized with appropriate synthesis of multiple data sources. Practical utility (weight = 0.15) evaluates whether the information provided is actionable for environmental decision-making.

Each criterion was scored on a 1--5 scale following detailed rubrics provided to the judge model. The weighted score was computed as:
\begin{equation}
S_{\text{weighted}} = 0.25 \cdot G + 0.25 \cdot A + 0.20 \cdot C + 0.15 \cdot H + 0.15 \cdot U
\end{equation}
where $G$, $A$, $C$, $H$, and $U$ denote grounding, scientific accuracy, completeness, coherence, and practical utility scores, respectively.

\subsubsection{Judge Protocol} \label{sec:judge_protocol}

For each evaluation cycle, the Judge LLM received a structured prompt containing the original user query, location metadata (coordinates, year, intent type), and the system's complete response. The prompt included the evaluation rubrics and instructed the judge to provide scores for each criterion along with brief reasoning. To assess inter-judge reliability, we examined score distributions across the four judge models and computed pairwise correlations of weighted scores for overlapping query--response pairs.


\section{Results} \label{sec:results}

In this section, we present findings organized around the four research questions posed in Section~\ref{sec:introduction}. We first examine whether AlphaEarth embedding dimensions encode physically meaningful environmental features (RQ1), then assess the spatial and temporal robustness of these relationships (RQ2). We describe how the validated interpretations enable a Land Surface Intelligence system (RQ3), and finally evaluate the system's response quality through an LLM-as-Judge framework (RQ4).

\subsection{RQ1: Do AlphaEarth Embeddings Encode Physically Meaningful Environmental Features?} \label{sec:results_rq1}

Our first research question asks whether the 64-dimensional AlphaEarth embedding space encodes identifiable relationships with physical land surface properties. To answer this, we break it down into four parts: (1) what linear associations exist between individual dimensions and environmental variables, (2) how much variance in environmental variables can the full embedding space explain through nonlinear models, (3) what additional structure does a jointly trained Transformer reveal, and (4) do the three analytical methods converge on consistent dimension--variable assignments.

\subsubsection{What linear associations exist between dimensions and environmental variables?} \label{sec:results_spearman}

\begin{figure*}[!ht]
\centering
\includegraphics[width=\textwidth]{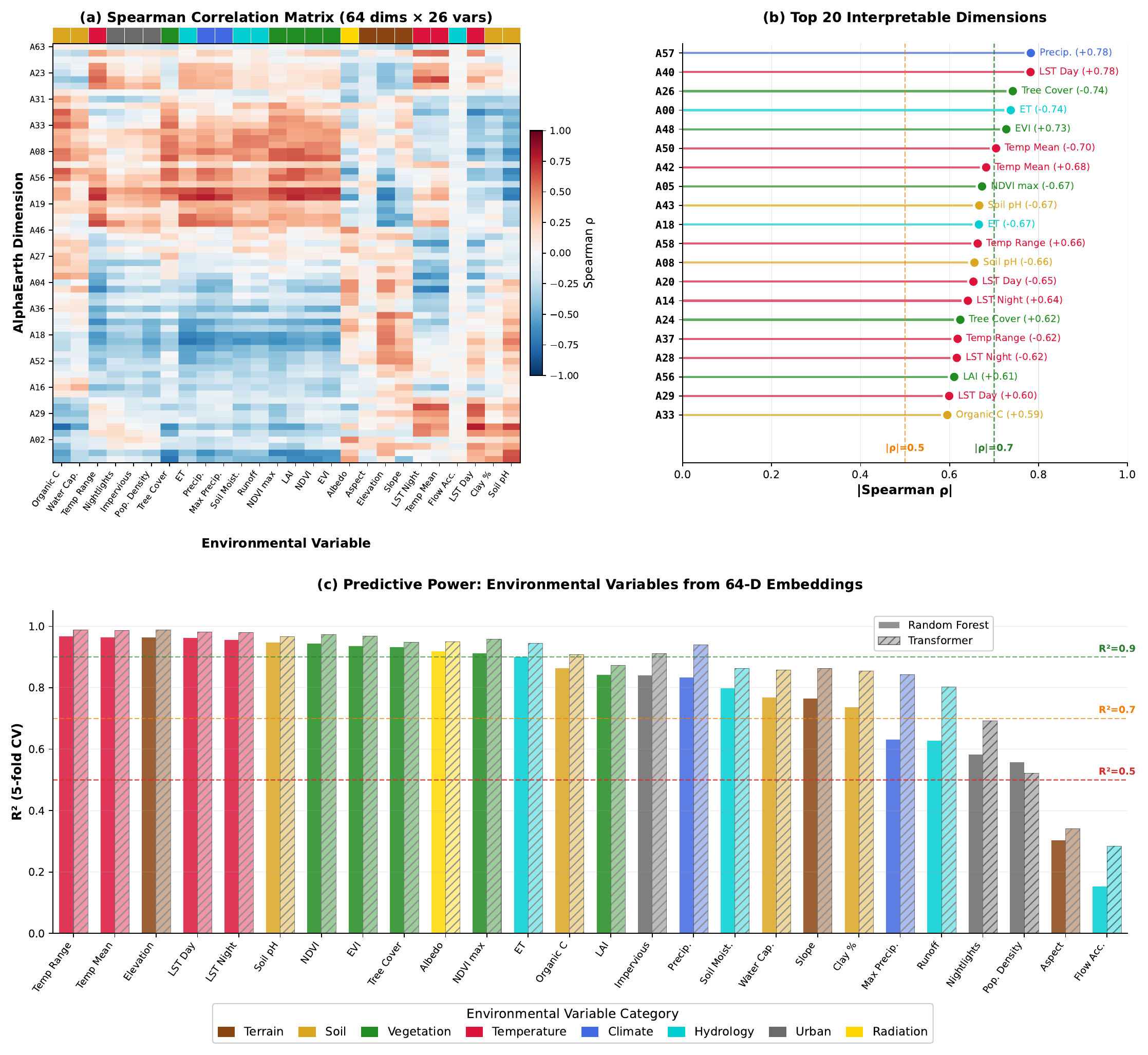}
\caption{AlphaEarth embedding interpretability analysis. (a) Spearman rank correlation matrix ($64 \times 26$) between embedding dimensions and environmental variables, with hierarchical clustering applied to both rows and columns. Color bars along the top axis indicate the thematic category of each environmental variable. (b) The 20 most interpretable dimensions ranked by their strongest absolute Spearman correlation ($|\rho|$), with labels indicating the primary associated variable. Vertical reference lines mark $|\rho| = 0.5$ and $|\rho| = 0.7$. (c) Predictive power ($R^2$, 5-fold cross-validation) of Random Forest and Transformer models for each environmental variable, using the 64-dimensional embedding as input. Variables are sorted by decreasing $R^2$; horizontal reference lines mark $R^2 = 0.5$, 0.7, and 0.9.}
\label{fig:interpretability}
\end{figure*}

The $64 \times 26$ Spearman correlation matrix reveals structured, physically interpretable relationships across the embedding space. Figure~\ref{fig:interpretability}a shows the full matrix after hierarchical clustering, and Figure~\ref{fig:interpretability}b ranks the 20 most interpretable dimensions by their strongest absolute correlation.

Of the 64 dimensions, 34 exhibit at least one correlation exceeding $|\rho| > 0.5$, 62 of 64 exceed $|\rho| > 0.3$, and six dimensions surpass $|\rho| > 0.7$. The strongest correlations span distinct environmental domains. Dimension A57 correlates most strongly with annual precipitation ($\rho = +0.78$) and also shows substantial associations with leaf area index ($\rho = +0.76$), dew point temperature ($\rho = +0.74$), and elevation ($\rho = -0.70$). This multi-variable profile reflects the coupling between orographic precipitation, vegetation productivity, and temperature modulation in complex terrain \citep{daly2008physiographically}. Dimension A40 captures thermal regime, with correlations to daytime LST ($\rho = +0.78$) and an inverse relationship with soil organic carbon ($\rho = -0.78$), consistent with faster decomposition rates in warmer environments \citep{davidson2006temperature}. Dimension A48 encodes vegetation structure: EVI ($\rho = +0.73$), LAI ($\rho = +0.72$), and NDVI ($\rho = +0.71$). Dimension A26 captures forest cover ($\rho = -0.74$ with tree cover), A00 encodes hydrological cycling (evapotranspiration $\rho = -0.74$, precipitation $\rho = -0.65$, soil moisture $\rho = -0.61$), and A50 tracks mean air temperature ($\rho = -0.70$).

Figure~\ref{fig:interpretability}a also shows a block-diagonal structure emerging from the hierarchical clustering. Dimensions group by thematic similarity: temperature-related dimensions cluster together, as do vegetation and hydrology dimensions. This suggests that the AlphaEarth training process organizes the embedding space along physically coherent axes rather than distributing environmental information arbitrarily. The spatial maps in Figure~\ref{fig:spatial_interp} further confirm this, showing that the geographic distribution of individual embedding dimensions closely mirrors the spatial patterns of their most correlated environmental variables.

\subsubsection{How much variance can the full embedding space explain?} \label{sec:results_rf}

Random Forest regression quantifies how well all 64 dimensions collectively predict each environmental variable. Figure~\ref{fig:interpretability}c compares RF and Transformer $R^2$ values across all 26 variables.

Under 5-fold cross-validation, 12 of 26 variables achieve $R^2 > 0.9$ and 20 of 26 exceed $R^2 > 0.7$. The strongest predictions are for temperature-related variables: dew point temperature ($R^2 = 0.97$), mean air temperature ($R^2 = 0.97$), and daytime LST ($R^2 = 0.96$). Elevation ($R^2 = 0.96$) and nighttime LST ($R^2 = 0.96$) are also near-perfectly reconstructed. Among vegetation indices, NDVI ($R^2 = 0.94$), EVI ($R^2 = 0.94$), tree cover ($R^2 = 0.93$), and albedo ($R^2 = 0.92$) are well captured. Evapotranspiration ($R^2 = 0.90$) and soil moisture ($R^2 = 0.80$) show strong but lower predictive power. Urban indicators, including nightlights ($R^2 = 0.58$) and population density ($R^2 = 0.56$), are more weakly encoded.

Permutation importance identifies which dimensions drive these predictions. For vegetation variables, A57 and A48 consistently rank among the top three predictors (NDVI: A57 first, A48 second; EVI: A57 first, A48 second). Temperature predictions rely on a distinct set: daytime LST depends primarily on A40, while nighttime LST and mean air temperature rely on A14 and A50. This separation indicates that the embedding space distinguishes between daytime radiative heating and nighttime thermal inertia. The least predictable variables, aspect ($R^2 = 0.30$) and flow accumulation ($R^2 = 0.15$), likely reflect the limited ability of 1~km-scale satellite embeddings to resolve local hillslope orientation and upstream catchment geometry.

\subsubsection{What additional structure does the Transformer reveal?} \label{sec:results_transformer}

The multi-task Transformer, jointly predicting all 26 variables from the 64-dimensional input, achieves higher $R^2$ than Random Forest for all 26 variables (Figure~\ref{fig:interpretability}c). Fourteen variables exceed $R^2 > 0.9$ under random cross-validation (versus 12 for RF), and 20 exceed $R^2 > 0.8$. The largest improvements appear for precipitation-related variables: annual precipitation increases from $R^2 = 0.83$ (RF) to $R^2 = 0.92$ (Transformer), and maximum monthly precipitation from $R^2 = 0.63$ to $R^2 = 0.78$. Dew point temperature ($R^2 = 0.99$) and mean air temperature ($R^2 = 0.98$) approach near-perfect prediction.

The Transformer's self-attention mechanism reveals inter-dimension relationships not captured by the other two methods. Attention weights from the final encoder layer, averaged across 200,000 forward passes, show that dimension A16 functions as an attention hub, receiving disproportionately high attention from most other dimensions. This suggests A16 encodes a broadly relevant conditioning feature, potentially related to geographic or climatic context, that the model uses when predicting across multiple target variables.

\subsubsection{Do the three methods converge?} \label{sec:results_convergence}

\begin{figure*}[!ht]
\centering
\includegraphics[width=\textwidth]{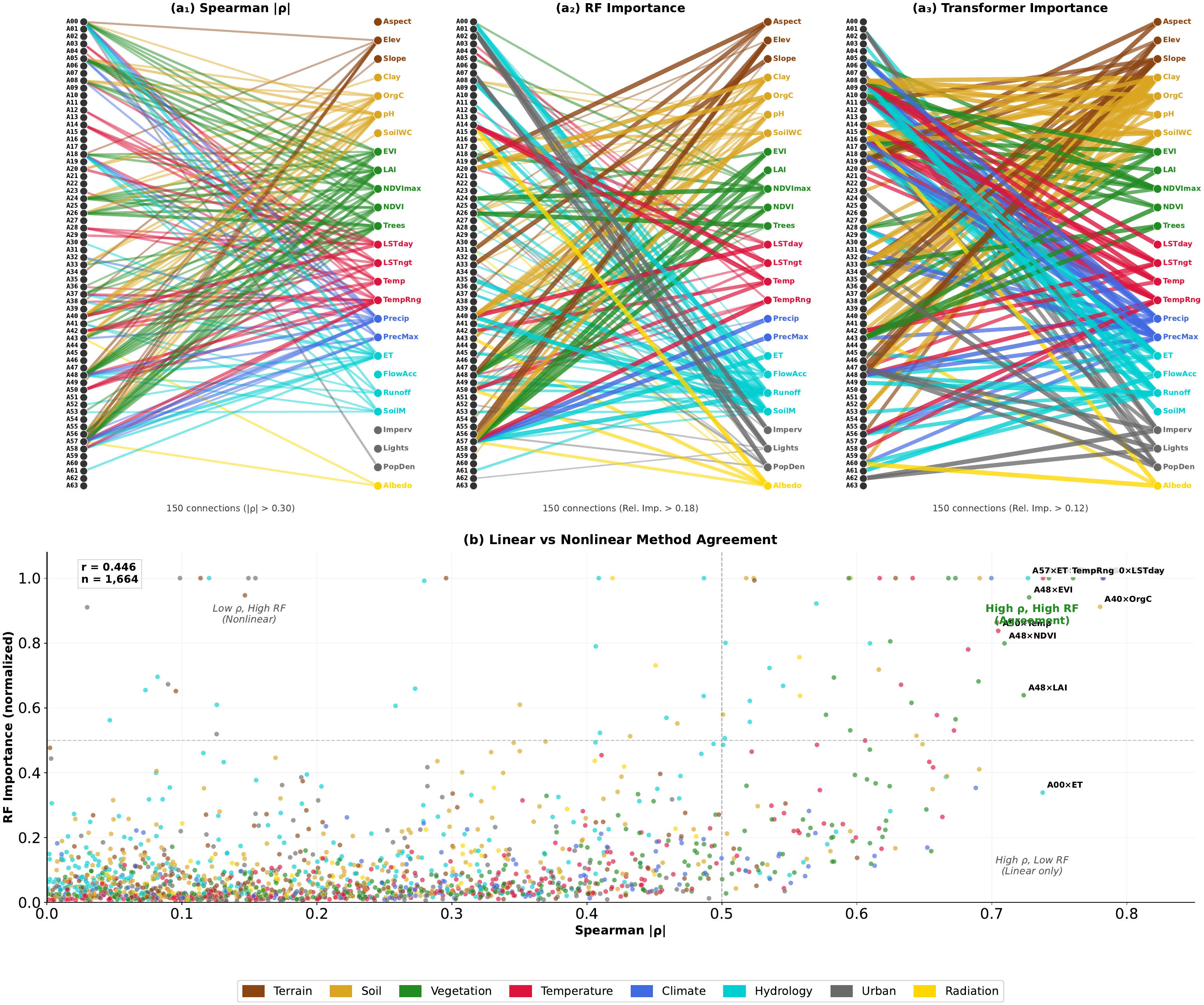}
\caption{Method convergence analysis. (a) Bipartite network graphs showing the primary dimension--variable connections identified by each of the three interpretability methods (Spearman, Random Forest, Transformer). Nodes on the left represent embedding dimensions and nodes on the right represent environmental variables, colored by thematic category. (b) Method agreement at the pair level, comparing absolute Spearman $|\rho|$ against Random Forest permutation importance across all $64 \times 26 = 1{,}664$ dimension--variable pairs.}
\label{fig:method_networks}
\end{figure*}

The three methods provide complementary rather than redundant characterizations of the embedding space. Figure~\ref{fig:method_networks} visualizes the dimension--variable mappings from each method as bipartite networks and quantifies their agreement.

When comparing primary variable assignments, Spearman and Random Forest agree for 11 of 64 dimensions (17.2\% two-way concordance). No dimension achieves three-way concordance across all methods, which is expected given that each method captures different aspects: Spearman measures monotonic trends, RF captures nonlinear importance through permutation, and the Transformer extracts gradient sensitivity within a jointly trained model.

The concordant dimensions are notable for their interpretive clarity. A57 is assigned to precipitation by both Spearman ($\rho = +0.78$) and RF (highest permutation importance). A40 maps to daytime LST, A26 to tree cover, and A48 to EVI under both linear and nonlinear analysis. Several temperature-encoding dimensions also agree: A50 and A42 map to mean air temperature, while A14 and A28 map to nighttime LST. These concordant dimensions tend to have the strongest absolute correlations (Figure~\ref{fig:interpretability}b), suggesting that the most prominent embedding--environment relationships are robust to analytical method.

Across all $64 \times 26 = 1{,}664$ dimension--variable pairs, the Pearson correlation between absolute Spearman $|\rho|$ and RF permutation importance is $r = 0.45$. Divergent dimensions, where methods disagree, tend to have weaker correlations where subtle nonlinear effects can shift assignments depending on the analytical lens. At the category level, Spearman assigns 26 of 64 dimensions primarily to temperature, while RF distributes assignments more broadly across urban (14), hydrology (12), and terrain (12). This difference arises because Spearman captures the dominant linear trend, which is often temperature-related, while RF detects complementary nonlinear contributions.

\subsection{RQ2: Are Embedding--Variable Relationships Spatially Robust and Temporally Stable?} \label{sec:results_rq2}

Our second research question addresses whether the dimension--variable relationships identified in RQ1 generalize across space and persist over time. We examine this in two parts: (1) spatial generalization assessed through block cross-validation, and (2) temporal stability across seven years.

\subsubsection{Do relationships hold under spatial validation?} \label{sec:results_spatial}

Spatial block cross-validation, using $2^{\circ} \times 2^{\circ}$ blocks as grouping units, tests whether predictive models exploit spatial autocorrelation or capture genuine physical structure. Figure~\ref{fig:validation}a shows the generalization gap ($\Delta R^2 = R^2_{\text{random}} - R^2_{\text{spatial}}$) for each variable.

For the Transformer, the mean gap across all 26 variables is $\Delta R^2 = 0.017$, and 18 of 26 variables exhibit gaps below 0.02. The smallest gaps occur for elevation ($\Delta R^2 = 0.002$), NDVI ($\Delta R^2 = 0.002$), and EVI ($\Delta R^2 = 0.002$), indicating that these relationships generalize well across regions. The largest gaps appear for runoff ($\Delta R^2 = 0.083$), maximum monthly precipitation ($\Delta R^2 = 0.066$), and clay fraction ($\Delta R^2 = 0.046$). These variables have strong regional heterogeneity and local determinants that may not transfer across spatial blocks.

For the Random Forest models evaluated on the top 10 variables, the mean gap is $\Delta R^2 = 0.009$, and all ten variables retain spatial $R^2 > 0.83$. The smallest RF gap is for precipitation ($\Delta R^2 = 0.002$) and the largest for soil organic carbon ($\Delta R^2 = 0.030$). The generally narrow gaps suggest that the embeddings encode physical structure rather than exploiting spatial proximity in the training data. This contrasts with prior reports of substantial spatial inflation in geospatial machine learning \citep{ploton2020spatial}, likely because AlphaEarth embeddings represent broad land surface characteristics that transcend local spatial patterns.

\subsubsection{Are relationships stable across years?} \label{sec:results_temporal}

\begin{figure*}[!ht]
\centering
\includegraphics[width=\textwidth]{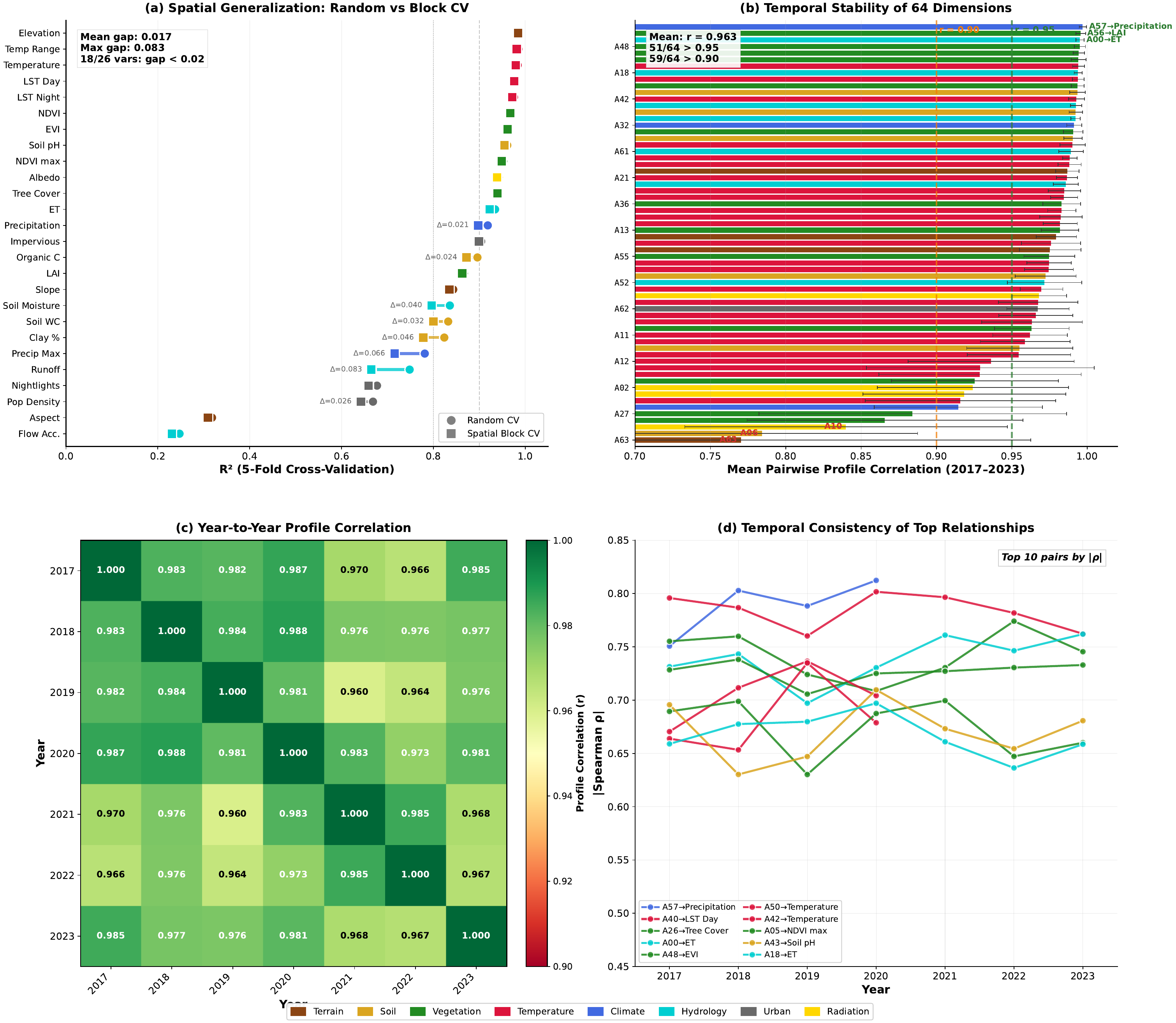}
\caption{Validation analysis. (a) Spatial generalization: comparison of random and spatial block cross-validation $R^2$ for each environmental variable using the Transformer model, with generalization gaps ($\Delta R^2$) annotated for variables where the gap exceeds 0.02. (b) Temporal stability of all 64 embedding dimensions, measured as mean pairwise Pearson correlation between annual Spearman profiles (2017--2023). Bars are colored by primary thematic category; vertical reference lines mark $r = 0.90$ and $r = 0.95$. (c) Year-to-year profile correlation matrix across the seven study years. (d) Temporal evolution of the top 10 dimension--variable pairs by absolute Spearman $|\rho|$ over 2017--2023.}
\label{fig:validation}
\end{figure*}

To assess temporal stability, we computed year-specific Spearman profiles ($n = 300{,}000$ samples per year, 2017--2023) and measured the mean pairwise correlation between each dimension's annual profiles. Figure~\ref{fig:validation}b shows the stability of all 64 dimensions, Figure~\ref{fig:validation}c shows the year-to-year correlation matrix, and Figure~\ref{fig:validation}d tracks the top dimension--variable pairs over time.

The mean temporal stability across all 64 dimensions is $\bar{r} = 0.963$. Of the 64 dimensions, 51 exceed $r > 0.95$ and 59 exceed $r > 0.90$ (Figure~\ref{fig:validation}b). Only two dimensions fall below $r = 0.80$: A63 ($r = 0.77$) and A06 ($r = 0.78$). These may encode features more sensitive to interannual variability or sensor-specific artifacts.

The year-to-year profile correlation matrix (Figure~\ref{fig:validation}c) shows uniformly high agreement, with values ranging from $r = 0.960$ (2019--2021) to $r = 0.988$ (2018--2020). No individual year appears as a systematic outlier. The most temporally stable dimensions include A57 ($r = 0.997$), which encodes precipitation, and A00 ($r = 0.995$), the hydrological cycling dimension.

Figure~\ref{fig:validation}d tracks the top 10 dimension--variable pairs across all seven years. The A57--precipitation relationship maintains $|\rho|$ values between approximately 0.76 and 0.82, while A40--LST Day fluctuates between approximately 0.76 and 0.80. These narrow ranges confirm that the physical interpretations assigned to each dimension are not artifacts of any single year's data but reflect stable properties of the underlying land surface.

\subsection{RQ3: Can Validated Interpretations Enable Retrieval-Augmented Generation?} \label{sec:results_rq3}

Our third research question asks whether the dimension--variable relationships established in RQ1 and validated in RQ2 can serve as the foundation for a natural language environmental query system. The Land Surface Intelligence system (Figure~\ref{fig:system_architecture}) operationalizes these relationships through three components: a dimension dictionary that translates embedding values into environmental meaning, a FAISS-indexed vector database that enables similarity search over the 12.1 million samples, and an LLM integration layer that generates natural language responses grounded in the retrieved data.

The dimension dictionary compiles the outputs of all three interpretability methods into a structured lookup. For dimensions with $|\rho| > 0.5$ (34 of the 64 dimensions), the system generates environmental interpretations at query time. For example, a high value in A57 combined with its positive correlation with precipitation ($\rho = +0.78$) is interpreted as indicating a wet climate regime; a high value in A48 combined with its EVI correlation ($\rho = +0.73$) signals high vegetation productivity. These interpretations are provided as structured context to the LLM alongside the actual environmental variable values retrieved from the metadata store.

\begin{figure*}[!ht]
\centering
\includegraphics[width=0.7\textwidth]{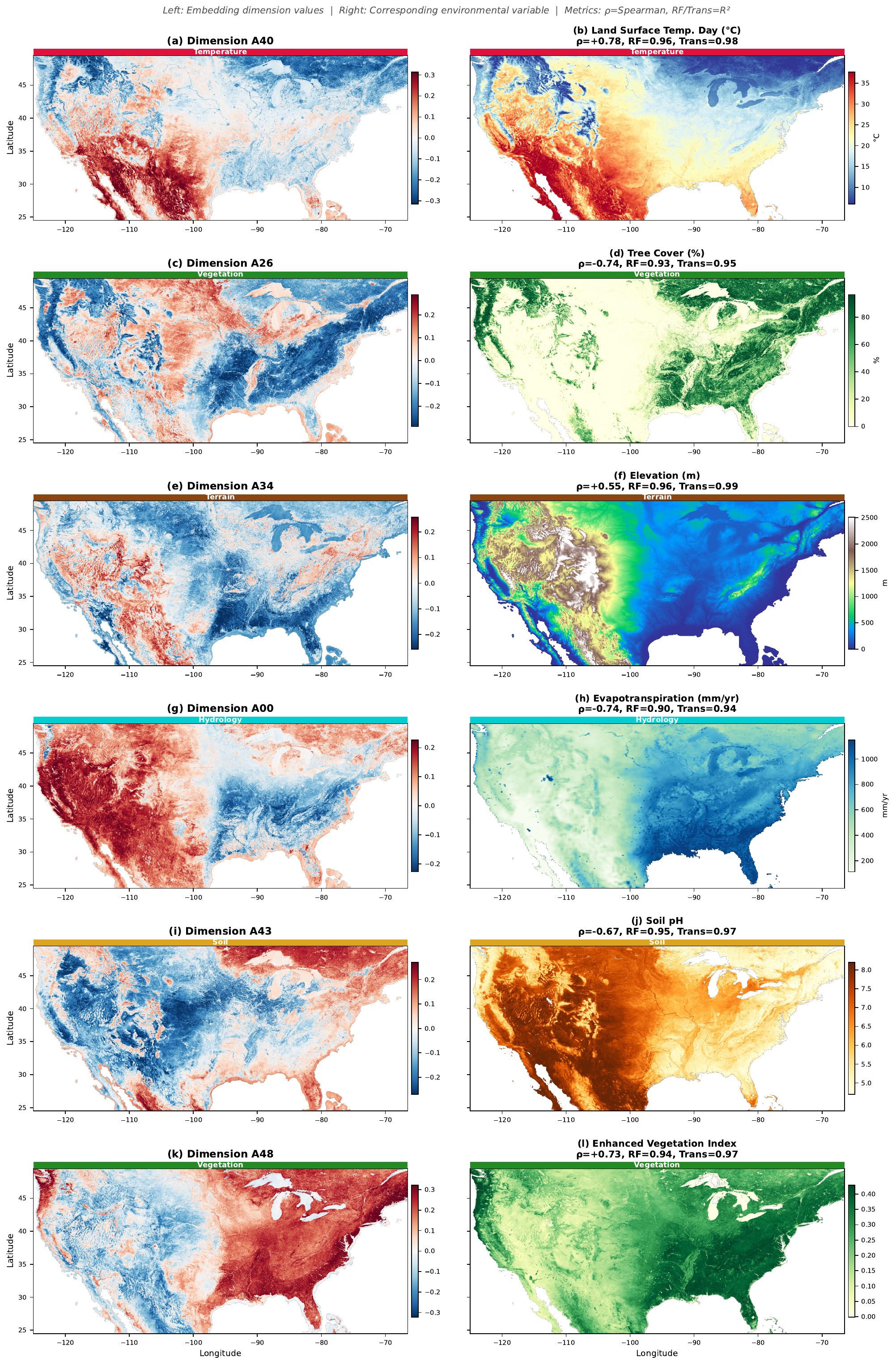}
\caption{Spatial interpretability demonstration. Six selected dimension--variable pairs spanning distinct environmental categories are shown. For each pair, the left panel maps the embedding dimension value and the right panel maps the corresponding environmental variable across CONUS. Spearman $\rho$, Random Forest $R^2$, and Transformer $R^2$ are reported for each pair. The spatial correspondence between embedding dimensions and their associated environmental variables confirms that the learned representations encode geographically coherent physical features.}
\label{fig:spatial_interp}
\end{figure*}

Similarity search over the FAISS index leverages a core property of the embedding space: proximity corresponds to physical similarity on the land surface. A query location in New England can retrieve analogous locations across CONUS, potentially in the Pacific Northwest or Appalachian foothills, enabling comparative environmental analysis without explicit feature matching. The spatial correspondence between embedding dimensions and environmental variables, visible in Figure~\ref{fig:spatial_interp}, underpins this retrieval mechanism.

Query intent classification shapes which variables and dimension interpretations are emphasized. Ten intent categories, ranging from flood risk and drought vulnerability to vegetation health and location profiling, each map to a prioritized set of environmental variables. This mapping is informed by the interpretability analysis: flood risk queries emphasize dimensions encoding precipitation (A57), soil moisture (A00), and runoff, while vegetation queries prioritize A48 (EVI) and A26 (tree cover). The alignment between intent categories and well-characterized dimensions is what enables grounded, domain-specific responses rather than generic environmental summaries.

\subsection{RQ4: How Can We Evaluate LLM-Based Geospatial Systems?} \label{sec:results_rq4}

Our fourth research question addresses the evaluation of LLM-based geospatial systems, which requires metrics beyond standard retrieval accuracy. We examine this in three parts: (1) overall system performance across 360 evaluation cycles, (2) variation across LLM backends, and (3) variation across query intent types.

\subsubsection{What is the overall response quality?} \label{sec:results_eval_overall}

\begin{figure*}[!ht]
\centering
\includegraphics[width=\textwidth]{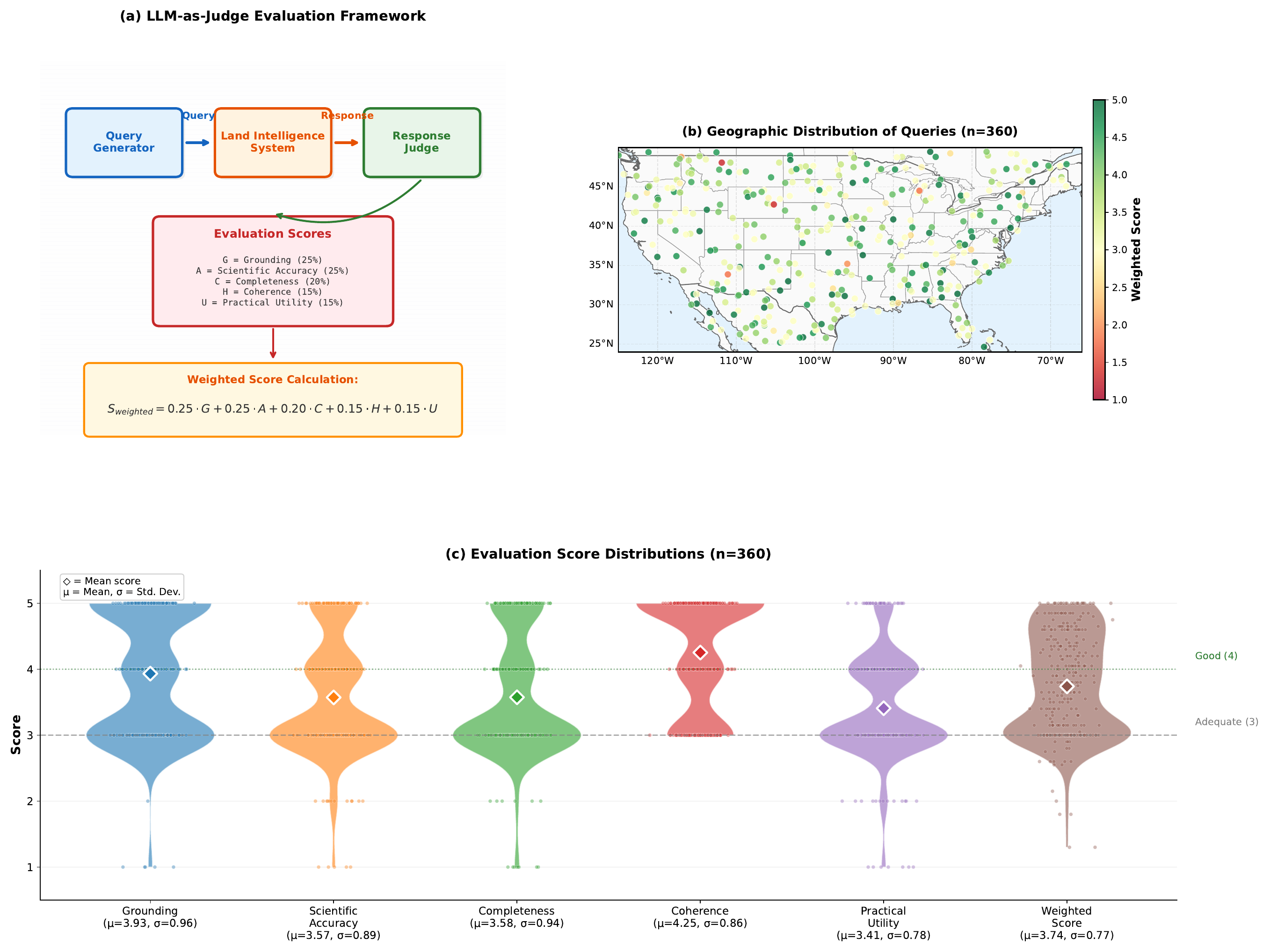}
\caption{LLM-as-Judge evaluation framework and results. (a) Schematic of the cross-model evaluation design: a Query Generator LLM produces natural language queries, the Land Intelligence System processes them with a designated system LLM, and a Judge LLM evaluates response quality across five weighted criteria. (b) Geographic distribution of the 360 evaluation queries across CONUS, colored by weighted score. (c) Score distributions for each evaluation criterion and the overall weighted score ($n = 360$). Diamond markers indicate the mean; $\mu$ and $\sigma$ are annotated below each criterion.}
\label{fig:eval_framework}
\end{figure*}

The cross-model evaluation experiment, comprising 360 query--response cycles across 12 configurations of four LLMs in rotating generator, system, and judge roles, yields a mean weighted score of $\mu = 3.74$ ($\sigma = 0.77$) on the 1--5 scale. Figure~\ref{fig:eval_framework}c shows the score distributions for each criterion.

All five criteria exceed the adequate threshold of 3.0. Coherence achieves the highest mean ($\mu = 4.25$, $\sigma = 0.86$), indicating well-structured response organization. Grounding ($\mu = 3.93$, $\sigma = 0.96$) confirms that responses frequently reference specific embedding dimensions and environmental variable values rather than producing generic text. Scientific accuracy ($\mu = 3.57$, $\sigma = 0.89$) and completeness ($\mu = 3.58$, $\sigma = 0.94$) fall in the adequate-to-good range. Practical utility receives the lowest mean ($\mu = 3.41$, $\sigma = 0.78$), suggesting room for improvement in translating environmental data into actionable guidance.

\begin{figure*}[!ht]
\centering
\includegraphics[width=\textwidth]{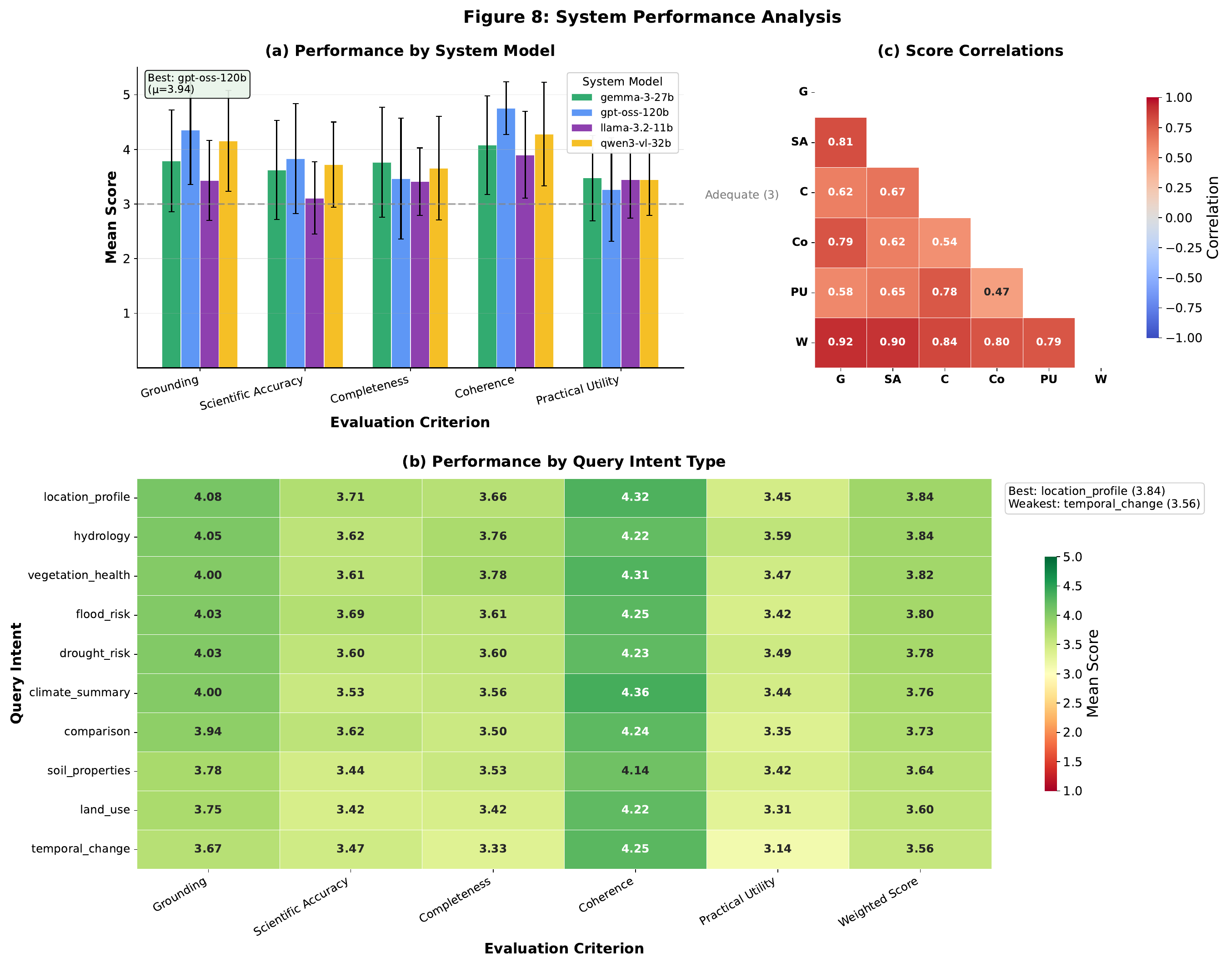}
\caption{System performance analysis. (a) Mean scores by evaluation criterion for each of the four system LLM backends. Error bars show one standard deviation. (b) Mean scores by query intent type across all evaluation criteria, with intents ranked by weighted score. (c) Pairwise Pearson correlations between the five evaluation criteria and the weighted score.}
\label{fig:performance}
\end{figure*}

Figure~\ref{fig:performance}c shows the inter-criteria correlation structure. The weighted score is most strongly driven by grounding ($r = 0.92$) and scientific accuracy ($r = 0.90$), confirming that the evaluation framework prioritizes data-driven responses. Grounding and coherence correlate at $r = 0.79$, suggesting that well-grounded responses tend to be better organized. Completeness and practical utility correlate at $r = 0.78$, indicating that more comprehensive responses are perceived as more useful. The weakest inter-criteria correlation is between coherence and practical utility ($r = 0.47$); a coherent response is not necessarily actionable without domain-specific recommendations.

\subsubsection{Does performance vary across LLM backends?} \label{sec:results_eval_models}

Figure~\ref{fig:performance}a compares performance across the four system models. GPT-OSS-120B achieves the highest weighted score ($\mu = 3.94$), with the advantage most pronounced for grounding and scientific accuracy. The larger reasoning-focused model more consistently references specific embedding data and environmental variables. Coherence scores are uniformly high across all models ($\mu > 4.0$), suggesting that even smaller models produce well-organized text when given structured context.

\subsubsection{Does performance vary by query intent?} \label{sec:results_eval_intents}

Figure~\ref{fig:performance}b shows systematic performance differences across intent types. Location profiling ($\mu = 3.84$) and hydrology ($\mu = 3.84$) achieve the highest weighted scores, followed by vegetation health ($\mu = 3.82$) and flood risk ($\mu = 3.80$). These intents align with the strongest embedding--variable relationships from our interpretability analysis: hydrology queries benefit from the strong encoding of precipitation (A57, $\rho = +0.78$), evapotranspiration (A00, $\rho = -0.74$), and soil moisture; vegetation queries leverage A48--EVI ($\rho = +0.73$) and A26--tree cover ($\rho = -0.74$).

Temporal change queries receive the lowest weighted scores ($\mu = 3.56$), reflecting the difficulty of assessing environmental change from single annual embedding composites without explicit multi-year comparison. Soil properties ($\mu = 3.64$) and land use ($\mu = 3.60$) also perform below average, consistent with the weaker encoding of soil and urban variables observed in the interpretability analysis ($R^2 < 0.80$ for clay fraction, nightlights, and population density).

This alignment between interpretability strength and query performance provides indirect validation of the system design. Intents mapping onto well-characterized dimensions receive better-grounded, more scientifically accurate responses, while intents requiring information from weakly encoded variables perform comparatively worse. Improving the embedding space's representation of soil and urban features would likely yield the largest gains for those query types.


\section{Conclusion} \label{sec:conclusion}

This study presents a systematic investigation of Google AlphaEarth satellite foundation model embeddings and their application to environmental information systems. Using 12.1 million samples across the Continental United States over seven years, we characterized relationships between the 64-dimensional embedding space and 26 environmental variables through three complementary methods: Spearman rank correlation for linear associations, Random Forest regression for nonlinear predictive relationships, and a multi-task Transformer for joint prediction and attention-based analysis. We then validated these relationships through spatial block cross-validation and temporal stability analysis, compiled the results into a dimension dictionary, and used this interpretive foundation to build a Land Surface Intelligence system that answers natural language environmental queries through retrieval-augmented generation over a FAISS-indexed embedding database.

Our interpretability analysis demonstrates that AlphaEarth embeddings are not opaque feature vectors but encode structured, physically coherent information about the land surface. Individual dimensions map onto specific environmental properties spanning climate, vegetation, hydrology, temperature, and terrain, and the strongest of these relationships are consistent across all three analytical methods. The embedding space collectively explains the majority of variance in most environmental variables, with temperature, elevation, and vegetation indices reconstructed with particularly high fidelity. Importantly, these relationships hold up under spatial block cross-validation, with narrow generalization gaps for the large majority of variables, and they remain stable across all seven years of available data. The few dimensions and variables that show weaker performance or lower stability point to specific limitations in how the foundation model represents certain features, particularly fine-scale terrain properties and urban indicators.

Building on the validated interpretations, the Land Surface Intelligence system demonstrates that satellite foundation model embeddings can serve as the retrieval backbone for a geospatial RAG pipeline. Because proximity in embedding space corresponds to physical similarity on the land surface, the system can retrieve environmentally analogous locations across the study area in response to natural language queries. The LLM-as-Judge evaluation, designed with rotating model roles to mitigate single-model bias, shows that the system produces responses that are well-grounded in actual satellite-derived data and scientifically coherent. The correspondence between interpretability strength and query-level performance demonstrates the connection between the analytical and applied components of this work.

Several limitations should be acknowledged. Our analysis is restricted to the Continental United States and to the seven years of available AlphaEarth annual composites; extending to other geographic regions and longer temporal baselines would test the generality of these findings. The evaluation relies on LLM-based judges and our domain expertise, while the rotating-role design mitigates some bias, further fine-tuning of our LLMs and system can be achieved through techniques such as reinforcement learning and low-rank adaptation \citep{hu2022lora}. Future work could address these gaps by incorporating higher-resolution ancillary data, extending the framework to global coverage, integrating multi-year embeddings for change detection, and evaluating the system with multiple domain experts in applied environmental decision-making contexts. More broadly, the interpretability-first approach demonstrated here is not specific to AlphaEarth; the same analytical framework could be applied to characterize and operationalize embeddings from other geospatial foundation models as they become available. Coupling validated embedding interpretations with targeted climate hazard data could enable specialized decision support systems for flood, drought, and wildfire risk that go beyond general environmental profiling toward actionable, location-specific climate adaptation guidance.

\section*{Acknowledgements}

The author thanks Lora Leligdon, Steve Gaughan, and Daniel Chamberlain of Dartmouth Libraries for their support during this research. The Land Surface Intelligence system was built using the Dartmouth Chat API (chat.dartmouth.edu), which provided access to multiple large language model backends for system implementation and evaluation.

\section*{Code and Data Availability}

The analysis code and supplementary materials for this study are publicly available at 
doi.org/10.5281/zenodo.18566431. Supplementary materials include 30 representative 
query--response--judgment cycles from the LLM-as-Judge evaluation, stratified by intent 
type, score range, and system model. The Land Surface Intelligence system code is not 
included in the public repository. These components depend on the Dartmouth Chat API 
\citep{dartmouth_chat}, which requires institutional authentication credentials issued 
by Dartmouth College.

All input datasets used in this study are publicly available through Google Earth Engine. The extracted dataset of co-located samples is available from the corresponding author upon reasonable request.

\bibliographystyle{elsarticle-num-names} 
\bibliography{bibliography}

\end{document}